\documentclass[11pt]{article}

\usepackage[preprint]{acl}

\usepackage{times}
\usepackage{latexsym}

\usepackage[T1]{fontenc}

\usepackage[utf8]{inputenc}

\usepackage{microtype}

\usepackage{inconsolata}

\usepackage{graphicx}

\usepackage{array}
\newcolumntype{P}[1]{>{\raggedright\arraybackslash}p{#1}}

\usepackage{xcolor}
\usepackage{amssymb}

\usepackage{amsmath}

\definecolor{darkgreen}{RGB}{0,110,0}
\definecolor{darkred}{RGB}{170,0,0}
\definecolor{royalblue}{RGB}{0,60,200}

\newcommand{\yes}{\textbf{\textcolor{darkgreen}{\large$\bullet$}}}     
\newcommand{\no}{\textbf{\textcolor{darkred}{\large$\times$}}}        
\newcommand{\partialyes}{\textbf{\textcolor{royalblue}{\large$\circ$}}} 

\usepackage{booktabs}
\usepackage{rotating}

\title{The Need for a Socially-Grounded Persona Framework for User Simulation}

\author{
Pranav Narayanan Venkit \quad
Yu Li \quad
Yada Pruksachatkun \quad
Chien-Sheng Wu\\
Salesforce Research \\
Palo Alto, CA, USA \\
\texttt{\{pnarayananvenkit,yu.li,ypruksachatkun,wu.jason\}@salesforce.com}
}

\begin{document}
\maketitle

\begin{abstract}
Synthetic personas are widely used to condition large language models (LLMs) for social simulation, yet most personas are still constructed from coarse sociodemographic attributes or summaries. We revisit persona creation by introducing \textbf{SCOPE}, a socially grounded framework for persona construction and evaluation, built from a 141-item, two-hour sociopsychological protocol collected from 124 U.S.-based participants. Across seven models, we find that demographic-only personas are a structural bottleneck: demographics explain only $\sim$1.5\% of variance in human response similarity. Adding sociopsychological facets improves behavioral prediction and reduces over-accentuation, and non-demographic personas based on traits and identity achieve strong alignment with substantially lower bias. These trends generalize to SimBench (441 aligned questions), where SCOPE personas outperform default prompting and NVIDIA Nemotron personas, and SCOPE augmentation improves Nemotron-based personas. Our results indicate that persona quality depends on sociopsychological structure rather than demographic templates or summaries.
\end{abstract}

\section{Introduction}

\begin{figure*}[t]
    \centering
    \includegraphics[scale=0.24]{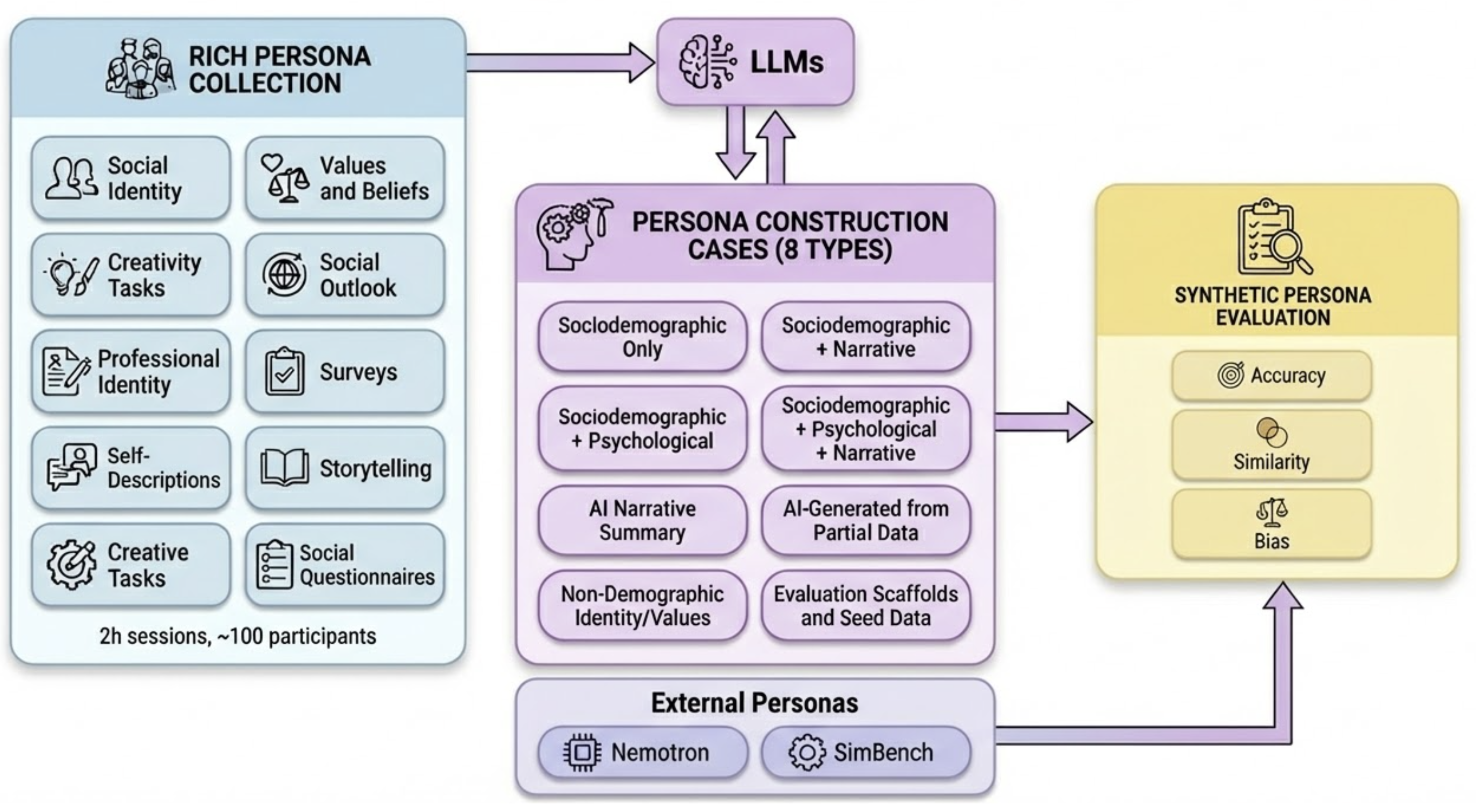}
    \caption{
        \textbf{End-to-end pipeline of our persona construction and evaluation framework.}
        The three major stages are: 
        (1) \textbf{Rich Persona Collection}: a two-hour, 141-item sociopsychological survey capturing eight  facets across 124 participants; 
        (2) \textbf{Persona Construction Cases}: structured persona representations generated by selectively combining demographic, behavioral, psychological, and narrative facets, as well as AI-generated summaries; 
        (3) \textbf{Synthetic Persona Evaluation}: measuring accuracy, similarity to human responses, and group bias. 
    }
    \label{fig:pipeline}
\end{figure*}

LLMs and agentic AI systems are increasingly deployed in settings where they must reason about, simulate, or stand in for human behavior \cite{batzner2025whose, gui2023challenge}. These include conversational assistants, recommender systems, safety and fairness evaluation, policy analysis, and agent-based social simulation \cite{mullick2024persona, benary2023leveraging, qiu2025mmpersuade}. To execute these tasks, recent work has adopted \textbf{synthetic personas}: \textit{structured representations of hypothetical individuals used to condition model outputs toward particular traits, identities, or behavioral tendencies} \cite{zhang2018personalizing, park2023generative}. Despite their growing importance, the construction of personas in NLP and AI has remained narrow and under-theorized \cite{gupta2023bias}. As summarized in Table~\ref{tab:persona_taxonomy}, most existing approaches rely on short free-text descriptions, small sets of sociodemographic attributes, or model-written summaries \cite{batzner2025whose, venkit2025tale}. Even large-scale persona resources such as PersonaHub \cite{ge2024scaling}, Nemotron \cite{nvidia/Nemotron-Personas-USA}, and related systems treat demographics or brief identity cues as proxies for human individuality \cite{kane2023we}. Table~\ref{tab:persona_comparison} further shows that these frameworks capture only limited slices of sociopsychological structure, typically omitting psychological traits, values, and behavioral patterns that social science research identify as central drivers of behavior \cite{turner2012self, costa2000neo}. This reliance on demographic-centric abstractions persists despite evidence that demography alone are weak predictors of individual-level behavior \cite{salminen2019detecting, khan2024mitigating}. 


We therefore revisit persona creation to address both gaps by introducing \textbf{SCOPE} (\textbf{S}ociopsychological \textbf{C}onstruct of \textbf{P}ersona \textbf{E}valuation), a human-grounded framework for constructing and evaluating synthetic personas. Motivated by the limitations of existing paradigms (Table~\ref{tab:persona_taxonomy}), SCOPE models personas as multidimensional sociopsychological profiles rather than demographic templates or narrative summaries. We collect rich persona data from 124 U.S.-based participants using a two-hour, 141-item protocol spanning \textbf{eight facets}: \textit{demographics, sociodemographic behavior, values and motivations, personality traits, behavioral patterns, identity narratives, professional identity}, and \textit{creativity}. 
Importantly, SCOPE also introduces an evaluation paradigm centered on \textit{structural fidelity}. Instead of asking whether a persona reproduces individual answers verbatim, we ask whether it reproduces the \textit{pattern} of human responses. We perform this using correlation-based alignment metrics, complemented by exact-match accuracy and a formal measure of demographic bias accentuation. 

Across seven model families\footnote{GPT4o, GPT5.1, Claude 3.5-Sonnet, Gemini 2.0 Flash, Gemini 2.5 Pro Thinking, DeepSeek R1, Qwen3} and a broad set of persona construction strategies, our results reveal a consistent pattern. First, demographic-only personas constitute a structural bottleneck: although demographic similarity explains only $\sim$1.5\% of behavioral variance among humans, demographic-only prompting often more than doubles this signal in model behavior, indicating systematic overgeneralization. Second, adding sociopsychological grounding steadily improves behavioral alignment while reducing demographic over-accentuation, with full SCOPE conditioning achieving the strongest overall correlation. Third, non-demographic personas based on values and identity alone can match or exceed fully conditioned personas while substantially reducing demographic bias. Finally, longer or more fluent model-written summaries do not substitute for structured, human-grounded facets, demonstrating that persona \textit{structure} matters as much as information quantity.
We further validate these findings on SimBench \cite{hu2025simbench}, a large external benchmark of social and behavioral questions, where SCOPE personas outperform default prompting and NVIDIA Nemotron personas, and SCOPE-based augmentation improves Nemotron’s performance. By grounding persona construction and evaluation in human data, SCOPE provides a foundation for building personas that are more behaviorally realistic, less demographically stereotyped, and better suited for sociotechnical NLP systems \cite{venkit2025deeptrace}. SCOPE also acts as an augmentation pipeline to existing frameworks making it immediately actionable for practitioners who rely on large-scale persona generation pipelines. 


\begin{table*}[t]
\centering
\footnotesize
\begin{tabular}{P{2.6cm} P{5.0cm} P{3.2cm} P{3.0cm}}
\hline
\textbf{Paradigm} & \textbf{Description} & \textbf{Captured Dimensions} & \textbf{Representative Example} \\ 
\hline

\textbf{Text-Based Descriptions} & 
Short free-text summaries describing personality, background, or preferences. & 
Low: narrative-only; limited demographic grounding. & 
\citet{ge2024scaling} \\

\textbf{Sociodemographic Inputs} & 
Personas constructed from attributes such as age, race, gender, etc & 
Low--medium: demographic-only. & 
\citet{nvidia/Nemotron-Personas-USA} \\

\textbf{Demographic + Identity Signals} & 
Adds identity questions, cultural background, or identity markers on top of demographic variables. &
Medium: demographics + self-identity cues. & 
\citet{lee2025spectrum} \\

\textbf{Interview-Based Personas} & 
Personas formed from interview transcripts or long-form questionnaire responses. &
Medium--high: narrative + behavioral profiles. & 
\citet{park2023generative} \\

\textbf{Our Framework} & 
Personas from sociopsychological data across eight facets. & 
Facet sociopsychological profiles. & 
\textit{This work} \\
\hline
\end{tabular}
\caption{Overview of major persona construction frameworks in current LLM research. Example of each of the framework is present in Table~\ref{tab:persona_examples}--\ref{tab:scope_case_persona_examples} in the Appendix.}
\label{tab:persona_taxonomy}
\end{table*}

\section{Related Work}
\label{sec:related}



\subsection{What Constitutes a Persona?}

In research fields of sociology and psychology, a persona is not reducible to a demographic profile or a short narrative description. Instead, human behavior is understood as emerging from interacting layers of social, psychological, and contextual structure. Social Identity Theory emphasizes that individuals derive meaning and norms from group memberships (e.g., race, gender, profession), but does not treat demographic categories as deterministic predictors of individual behavior \citep{tajfel2001integrative, turner2012self}. Personality psychology further demonstrates that stable traits, most prominently the Big Five to predict consistent patterns of cognition, affect, and behavior across contexts \citep{costa2000neo, costaMcCrae1992}. Complementing traits, value theories such as Schwartz’s universal values framework show that motivational priorities systematically shape moral judgments, preferences, and life choices as well \citep{schwartz1992universals}. Beyond traits and values, narrative identity theory shows that individuals organize their experiences through internalized life stories, which provide meaning, and a goal structure \citep{mcadams1995, mcadams2001psychology}. These narratives influence how people interpret situations and decisions, particularly in ambiguous or moral domains. Finally, person–situation interaction models emphasize that behavior is best when structured patterns across situations rather than single responses in isolation \citep{mischel1995cognitive}. 

\subsection{Persona Construction in NLP}

Recent work constructs personas as short textual profiles or self-descriptive sentences appended to dialogue, popularized by PersonaChat-style datasets and models \citep{zhang2018personalizing}. More recent persona resources extend this approach by generating personas from sociodemographic attributes, often aligned with census statistics. Examples include Persona Hub \citep{ge2024scaling} and NVIDIA’s Nemotron Persona \cite{nvidia/Nemotron-Personas-USA}, which provides millions of fully synthetic U.S. personas. While effective for scale, these approaches largely treat personas as demographic or lightly behavioral templates. Empirical analyses increasingly show that such representations risk collapsing behavioral diversity into demographic stereotypes \cite{venkit2025tale, cheng2023marked, guo2025exposing}. 
A growing body of work also argues that persona quality depends on psychologically meaningful structure. SPeCtrum models identity as a composition of social identity, personal identity, and life context \citep{lee2025spectrum}. PB\&J incorporates psychologically grounded rationales to improve preference and judgment prediction \citep{joshi2025improving}. ValueSim focuses on simulating individual value systems via structured backstories \citep{du2025valuesim}. However, with an increasingly diverse approach in persona creation, there is less clarity on what facets or approaches actually represent users. 



Synthetic personas are also increasingly used as virtual participants in social science, policy analysis, and system evaluation. SimBench aggregates large-scale social and behavioral datasets to benchmark whether LLMs can reproduce population-level response patterns \citep{hu2025simbench}. Other work evaluates LLMs as substitutes for human survey respondents, highlighting their promise and limitations \citep{argyle2023out, kolluri2025finetuning}. In recommender systems, persona-based modeling has been shown to improve preference prediction when grounded in psychologically meaningful features rather than demographics alone \citep{joshi2025improving, ketipov2023predicting}.




\begin{table*}[t]
\centering
\footnotesize
\begin{tabular}{p{4.3 cm} c c c c c}
\hline
\textbf{Persona Data Type / Facet} &
\textbf{Persona Hub} &
\textbf{Nemotron} &
\textbf{SPeCtrum} &
\textbf{Stanford 1000} &
\textbf{SCOPE (Ours)} \\
\hline

Demographic Information &
\yes & \yes & \yes & \yes & \yes \\

Sociodemographic Behavior &
\no & \no & \partialyes & \partialyes & \yes \\

Values \& Motivations &
\no & \no & \no & \partialyes & \yes \\

Personality Traits (Big Five) &
\no & \no & \no & \no & \yes \\

Behavioral Patterns \& Preferences &
\no & \no & \partialyes & \partialyes & \yes \\

Identity Narratives &
\yes & \partialyes & \yes & \yes & \yes \\

Professional Identity &
\no & \yes & \no & \partialyes & \yes \\

Creativity \& Cognitive Flexibility &
\no & \no & \no & \no & \yes \\

Grounding in Human Data &
\no & \no & \no & \yes & \yes \\

Sociopsychological Structure &
\no & \no & \no & \no & \yes \\

Designed for Evaluation &
\no & \yes & \no & \no & \yes \\
\hline
\end{tabular}

\vspace{2mm}
\textit{\yes\ = information present;\quad
\no\ = information absent;\quad
\partialyes\ = partially or implicitly supported.}

\caption{
Comparison of major persona construction approaches (Persona Hub \cite{ge2024scaling}, Nemotron \cite{nvidia/Nemotron-Personas-USA}, SPeCtrum \cite{lee2025spectrum} \& Standford 1000 \cite{park2024generative}).  
SCOPE is the only framework that integrates all eight sociopsychological facets and explicitly separates conditioning versus evaluation dimensions. 
}
\label{tab:persona_comparison}
\end{table*}

\section{Socially Grounded Persona Framework}

Synthetic personas are widely used in NLP and AI, yet their construction remains under-theorized and weakly grounded in empirical social science. Prior work in computational personality and behavior modeling shows that psychological traits and values are substantially more predictive of behavior than sociodemographic variables alone, a distinction largely absent from current persona construction practices \cite{ketipov2023predicting, joshi2025improving}.
To address this gap, we introduce \textbf{SCOPE: Sociopsychological Construct of Persona Evaluation}, a sociopsychologically informed framework for constructing rich, multidimensional persona. 


\subsection{The SCOPE Framework}

To move beyond demographic-only personas, SCOPE introduces a multidimensional sociopsychological architecture grounded in established findings from social psychology, personality science, narrative identity theory, and computational behavior modeling. 
SCOPE defines a structured persona representation consisting of \textbf{eight facets} (Table~\ref{tab:scope_facets}). Each facet is defined through structured questions and free-text inputs. SCOPE distinguishes between two categories of facets:

\noindent\textbf{Conditioning Facets} (used for persona construction).  
These include \emph{Demographics}, \emph{Sociodemographic Behavior}, \emph{Personality Traits}, and \emph{Identity Narratives}, which provide the social, psychological, and narrative grounding for persona generation.

\noindent\textbf{Evaluation Facets} (held out during construction).  
These include \emph{Values \& Motivations}, \emph{Behavioral Patterns}, \emph{Professional Identity}, and \emph{Creativity \& Innovation}, which serve as behavioral and attitudinal targets for evaluating persona coherence.
 

Each facet is grounded in established sociological and psychological frameworks. Table~\ref{tab:persona_taxonomy} details the content for each facet. This partitioning reflects a core principle of sociopsychological modeling: identity is partly observable (e.g., demographics and personality traits) and partly inferential (e.g., values, motivations, and creative expression). Accordingly, SCOPE provides both (1) a mechanism for constructing multidimensional LLM personas (2) a framework for evaluating their behavioral grounding. In total, SCOPE comprises \textbf{141 curated attributes} for all facets (Appendix~\ref{sec:appendix_dataset_stats}).

\subsection{Data Collection Design}

To establish a socially grounded foundation for SCOPE, we designed a two-hour survey capturing detailed sociopsychological data across the eight facets in Table~\ref{tab:scope_facets}. This enables direct comparison with real human responses, unlike prior work that relies primarily on fully synthetic data. The final instrument contains 141 items administered via Typeform\footnote{https://www.typeform.com/}
 and spans nominal, ordinal, and free-text formats. Participants completed \textit{51 Likert-scale items, 37 single-choice dropdown questions, 20 multiple-choice questions, 3 short free-text inputs,} and \textit{30 narrative or creative prompts} (e.g., life stories, identity reflections, and hypothetical reasoning tasks) (Appendix~\ref{sec:appendix_survey_info}-\ref{sec:appendix_dataset_stats}).

Each facet was informed by established social-science frameworks. Personality items followed the BFI-2-S, value items drew on Schwartz’s value theory, and behavioral items were aligned with large-scale surveys such as Pew Internet \& Technology, the General Social Survey (GSS), and the World Values Survey (WVS) \cite{davis1999general, pewInternet, inglehart2014wvs}. All items were written as an original composite survey rather than directly reproducing any single instrument. We recruited a U.S.-based sample guided by Census-like proportions, with explicit emphasis on racial diversity. Recruitment occurred in two stages: (1) outreach via Twitter/X, LinkedIn, and professional Slack groups; and (2) stratified sampling through a commercial research platform (User Interviews). The study protocol was approved by the institution’s Ethics Office.

To ensure authentic responses, we employed instructional deterrents and automated screening \cite{zhang2024generative, christoforou2024generative}. GPTZero was used as a coarse AI-authorship filter, supplemented by manual review \cite{brown2023gptzero, gptzero}. Participants were compensated \$50; 124/146 completed the survey with verified human-authored responses, yielding 17,484 total responses. The final sample comprised 33.0\% White (41), 19.3\% African American or Black (24), 20.1\% Hispanic or Latino (25), 14.5\% Asian (18), 7.2\% American Indian or Alaska Native (9), 2.4\% Native Hawaiian or Pacific Islander (3), and 2.2\% Other (4). This closely aligns with U.S. Census distributions while modestly oversampling underrepresented groups. Dataset statistics and question-type distributions are in Appendix~\ref{sec:appendix_dataset_stats}.

\subsection{Synthetic Persona Construction}

Given the human-grounded SCOPE profiles, we next construct synthetic personas that can be created in an LLM and systematically varied in representational richness. For each of the 124 participants, we serialize their SCOPE profile into a structured ``persona scaffold'' using only the conditioning facets (Demographic Information, Sociodemographic Behavior, Personality Traits, Identity Narratives, Table~\ref{tab:scope_facets}), while holding out the remaining facets as evaluation targets. We define facet-ablation variants that differ only in which parts of the human profile are exposed to the model:

\noindent\textbf{Case 0: No Personas.} Baseline case that contains no persona informaiton. This case shows the improvement of using personas.

\noindent\textbf{Case 1: Demography-Only.} Personas are constructed from basic sociodemographic attributes. This resembles how most works construct personas.
    
\noindent\textbf{Case 2: Demography + Narratives.} Adds personal and professional identity narratives on top of demographics.
    
\noindent\textbf{Case 3: Demography + Traits.} Demographics with Big Five traits derived from the BFI-2-S.
    
\noindent\textbf{Case 4: Full Conditioning (Demography + Traits + Narratives + Sociodemographic Behavior).} Uses all conditioning facets..

We additionally introduce three \textbf{LLM-centric} variants to mirror current practices in automated persona generation:

\noindent\textbf{Case 5: LLM Narrative Summaries.} The LLM receives a condensed persona summary generated \emph{by the model itself} from the full SCOPE profile. This compresses response into a single narrative persona as seen in Nemotron.

\noindent\textbf{Case 6: LLM Facet Completion.} The model is given only partial information (e.g., demographics or demographics+identity) and instructed to infer the missing facets. Multiple ablation of what information is used will create case 6 subsets.

\noindent\textbf{Case 7: Non-Demographic Personas.} Personas that omit demographics entirely (e.g., narrative-only or trait+narrative personas). These variants evaluate whether non-demographic cues alone can support behavioral simulation.

Each variant is paired with the same set of held-out SCOPE questions and external task. We show examples of each case in Table~\ref{tab:scope_case_persona_examples} in the Appendix.

\section{Evaluation Framework} \label{sec:evaluation}

\begin{figure*}[t]
    \centering
    \includegraphics[scale = 0.2]{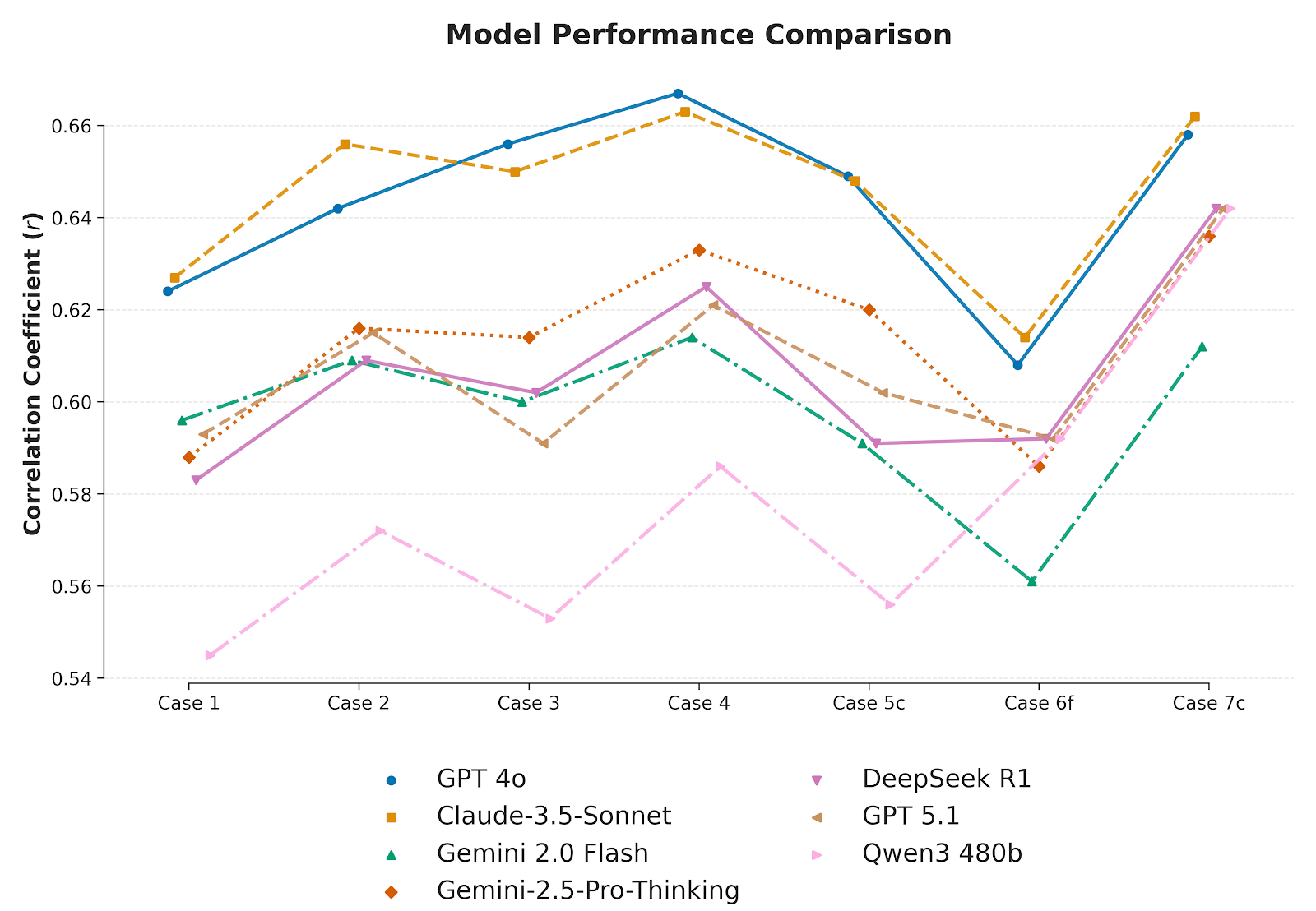}
    \caption{Line graph denotes the correlation score of all seven LLM modes. 
    }
    \label{fig:correlation_result}
\end{figure*}

Our evaluation intends to address a central question in persona modeling: rather than asking whether model responses \textit{match} human answers, we ask whether synthetic personas exhibit the \textit{same behavioral structure}. This distinction is essential as humans rarely provide identical responses, but they often display stable patterns across related questions \cite{kambhatla2022surfacing, venkit2025tale}. Given that humans themselves disagree item-by-item, we emphasize structure-preserving similarity: when a human answers relatively high on some questions and low on others, does the synthetic persona vary in parallel? We therefore evaluate personas along three axes:

\noindent\textbf{A. Behavioral correlation:} pattern similarity between human and model responses.\newline
\noindent\textbf{B. Accuracy:} exact agreement on social-attribute labels and Likert responses.\newline
\noindent\textbf{C. Bias:} extent to which a persona over- or under-accentuates demographic differences relative to real group responses. We look at this in Section~\ref{subsec:bias}

We first analyze these metrics on our SCOPE persona ablations (Cases~1--7), then evaluate generalization on SimBench \cite{hu2025simbench}, and finally study demographic bias.

\subsection{Baseline Persona Comparisons}
\label{subsec:baseline}

For each participant $i$, persona case $c$, and model $m$, let
$\mathbf{y}_i^{\text{human}} \in \mathbb{R}^K$ denote that participant's vector of held-out SCOPE answers (across $K$ evaluation questions), and $\mathbf{y}^{\text{model}}_{m,c,i} \in \mathbb{R}^K$ the answers generated by model $m$ when instantiated with case-$c$ persona information. We construct a golden standard text set consisting of questions from \textit{sociodemographic behaviours}, \textit{personal values and motivations}, and \textit{behavioral patterns} to understand how each case is able to predict and replicate these facets. This leads to a total MCQ set of 72 questions which is used as the first baseline of evaluation. 

\paragraph{Correlation.}
We measure behavioral alignment using the Pearson correlation coefficient
\begin{equation}
    \small
    r_{m,c,i} = \mathrm{corr}\!\big(
    \mathbf{y}^{\text{human}}_i,\;
    \mathbf{y}^{\text{model}}_{m,c,i}
    \big),
\end{equation}
and report the participant-level mean
\begin{equation}
    \small
    \bar{r}_{m,c} = \frac{1}{N} \sum_{i=1}^{N} r_{m,c,i}.
\end{equation}
Intuitively, $\bar{r}_{m,c}$ answers: \emph{does this persona reproduce the similar response characteristics as its human counterpart, up to a linear transformation?}

\paragraph{Accuracy.}
For completeness, we also compute an exact-match accuracy over all evaluation items:
\begin{equation}
    \small
    \mathrm{Acc}_{m,c} =
    \frac{1}{NK}
    \sum_{i=1}^{N} \sum_{k=1}^{K}
    \mathbf{1}\!\left[
    y^{\text{model}}_{m,c,i,k} = y^{\text{human}}_{i,k}
    \right],
\end{equation}
where equality is defined at the level of the discrete response option (Likert rating, multiple-choice option, or categorical label).

\paragraph{Do richer personas help?}
Figure \ref{fig:correlation_result} shows the correlation results across all the models used and Table~\ref{tab:case_gpt4} and \ref{tab:task_level_result} summarizes GPT-4o performance\footnote{We show GPT4o results as they are the better performing of all the models. The complete result is in the Appendix.} across the most informative persona cases; the full 7-model breakdown appears in Table~\ref{tab:full_results_models_a} -- \ref{tab:full_results_models_b}. Three trends emerge across all the models:

\noindent\textbf{I. Demographics alone are insufficient.} Demography-only personas (Case~1) achieve only moderate correlation ($\bar{r}=0.624$) and the lowest accuracy (35.1\%), despite being the dominant paradigm in current work.
    
\noindent\textbf{II. Adding sociopsychological facets improves alignment.} Sequentially enriching personas with identity narratives (Case~2), traits (Case~3), and full SCOPE conditioning (Case~4) steadily increases correlation and accuracy, with Case~4 reaching $\bar{r}=0.667$ and 39.7\% accuracy.
    
\noindent\textbf{III. Human-grounded facets outperform AI-compressed personas.} AI-generated narrative summaries (Case~5c) slightly improve over demographics but never surpass fully grounded representations, even when summaries are long. Non-demographic SCOPE personas built only from traits and idenitity narratives (Case~7c) match or exceed the best full-information case in correlation while avoiding direct use of demographic attributes.

\noindent\textbf{IV. Personas grounded in non-sociodemographic information perform better.} The greater finding show that to understand social behaviours, sociodemographic information is not the most important element of a persona. We see that for Case 6 iterations, where personas are created using either or both traits or identity narrative information tend to perform better, \textit{in regards to accuracy}, than the ones generated by sociodemographic information. 

\noindent\textbf{V. Socially grounded personas are able to simulate sociodemographic and behavioral patters.} Table \ref{tab:task_level_result} also shows how amongst all tasks, personas are able to replicate well sociodemographic behaviours and personality traits. The results also show that unique day to day behaviours are capturable but harder to replicate. The results show that personas can be used well for market specific understanding of a group pattern as compared to individualistic preferences. 
\begin{table}[t]
\centering
\small
\begin{tabular}{lccc}
\hline
\textbf{GPT-4o Persona Case} & $\bar{r}$ & Acc. & Bias \% \\
\hline
0: No Persona        & 0.448 & 20.47 & - \\
1: Demography only            & 0.624 & 35.07 & 101.23 \\
2: + Narratives      & 0.642 & 37.74 & \phantom{-}8.30 \\
3: + Traits    & 0.656 & 38.60 & 75.50 \\
4: Full conditioning          & 0.667 & 39.67 &  -6.40 \\
5c: LLM summary (1500w) & 0.649 & 38.07 & -38.42 \\
6f: AI narrative+traits        & 0.608 & 43.42 & -40.40 \\
7c: traits + narrative only    & 0.658 & 39.68 & -56.35 \\
\hline
\end{tabular}
\caption{
Behavioral alignment of GPT-4o personas across key SCOPE cases. 
$\bar{r}$ is mean Pearson correlation with human responses; Acc. is exact-match accuracy (\%). 
Bias\% is defined in Eq.~\eqref{eq:bias_percent}. 
Full results for all models(1--7a/b/c, 5a--d, 6a--f) are in the appendix.
}
\label{tab:case_gpt4}
\end{table}

\begin{table}[t]
\centering
\footnotesize
\begin{tabular}{lccc}
\toprule
\textbf{Case} &
\textbf{Socio.} &
\textbf{Behavioral} &
\textbf{Traits} \\
\midrule
Case 1  & 0.4084 & 0.2172 & 0.3356 \\
Case 2  & 0.4292 & 0.2206 & 0.3792 \\
Case 3  & 0.4299 & 0.2301 & 0.3972 \\
Case 4  & \textbf{0.4438} & \textbf{0.2383} & \textbf{0.4051} \\
Case 5a & 0.4287 & 0.2166 & 0.3858 \\
Case 5b & 0.4313 & 0.2213 & 0.3866 \\
Case 5c & 0.4280 & 0.2200 & 0.3897 \\
Case 5d & 0.4257 & 0.2247 & 0.3957 \\
\bottomrule
\end{tabular}
\caption{
A snippet of facet-level accuracy scores across persona construction cases for Socio.: \textit{sociodemographic behavior}, Behavioural: \textit{behavioral patterns}, and
Traits: \textit{personality traits}.
}
\label{tab:task_level_result}
\end{table}

Across models, we see similar trends: richer, multi-facet personas consistently yield higher correlations than demographic-only cases, and non-demographic SCOPE personas (Case~7c) match or exceed full conditioning while remaining more conservative with respect to demographic bias. We additionaly also explore open ended questions in Appendix~\ref{sec:creativity_case_effects} where we use framework defined by \citet{venkit2025tale, chakrabarty2024art} to evaluate writing style and diversity of text based answers. Our findings show how writing styles are very facet dependent and as a task is sensitive.

\subsection{Task Based Performance Analysis}\label{subsec:simbench}
To test whether our findings generalize beyond our internal questionnaire, we evaluate SCOPE personas on SimBench, a composite benchmark built from 20 large-scale social and behavioral datasets~(e.g., AfroBarometer, ESS, ISSP, LatinoBarometro, OpinionQA). This provides a task-based understanding of how our persona framework performs well in behavioral and social tasks. We align SimBench questionnaire items with our personas created to retain 441 questions whose wording, response scale, and sociotechnical context are compatible with the sociodemographic groups present in our collected data and personas.

We created GPT-4o personas\footnote{as GPT4o is the current high-performing model} for each human participant and persona case, then prompted the model to answer the \textbf{441 SimBench questions} in character. For each SimBench question $k$ we aggregate the human responses for the matching demographic group into an empirical distribution $p_k$, and treat the persona answer as a degenerate distribution $q_k$. We then compute \textbf{majority-vote accuracy}, i.e., whether the persona selects the most frequent human response for that group (within 3 turns).

We also compare against personas sampled from the NVIDIA Nemotron USA persona dataset \cite{nvidia/Nemotron-Personas-USA}, using their full textual persona descriptions and the same GPT-4o backend. Illustrated examples of Simbench and Nemotron are in Appendix~\ref{sec:appendix_external_formats}.
SCOPE personas that include sociopsychological facets (Cases~2--4 and 7c) outperform Nemotron personas across the 441 SimBench items, while demographic-only SCOPE personas perform comparably to or slightly better than Nemotron. This supports our central claim: \emph{one-dimensional sociodemographic personas are not sufficient even on external benchmarks, whereas multi-facet SCOPE personas does better to real-world tasks}. This finding is greatly also seen where Nemotron personas, augmented with SCOPE facets, becomes the best performing variant within the Nemotron persona cases. This shows that\textit{ SCOPE framework can be used as an augmentation of existing personas to increase their social and behavioral information and attributes to obtain strongly similar predictive behaviors}.

\begin{table}[t]
\centering
\footnotesize
\begin{tabular}{l l c}
\toprule
\textbf{Persona Source} & \textbf{Case Description} & \textbf{Accuracy} \\
\midrule

\textbf{SCOPE (Ours)} 
& 4. Full conditioning 
& \textbf{0.584} \\

& 2. Demo. + Narratives 
& 0.557 \\

& 6a. AI profile
& 0.553 \\
& 3. Demo. + Traits 
& 0.531 \\

& 1. Demographics only 
& 0.518 \\

& 5d. AI summary
& 0.494 \\

\midrule
\textbf{Nemotron} 
& \textbf{SCOPE augmented} 
& 0.533 \\

& Persona text only
& 0.496 \\

& All Data 
& 0.446 \\

\midrule
\textbf{SimBench} 
& Default personas
& 0.436 \\

\midrule
\textbf{Baseline} 
& No Persona
& 0.258 \\

\bottomrule
\end{tabular}
\caption{
Task-based evaluation on \textbf{SimBench}.We report accuracy using GPT-4o personas under different construction strategies. 
}
\label{tab:simbench_results}
\end{table}

\subsection{Bias and Behavioral Disparities}
\label{subsec:bias}

Finally, we assess whether persona prompting \emph{amplifies} or \emph{attenuates} demographic structure relative to human responses. Our aim is not to remove legitimate demographic signal, but to detect when personas \textit{over-accentuate} demographic differences.

\noindent\textbf{Human baseline:}
We estimate the predictive strength of demographics on behavioral similarity using $N_{\text{human}}=124$ participants, 13 demographic variables, and 134 evaluation questions. For each participant pair $(i,j)$, we compute demographic similarity and response similarity using cosine similarity over standardized feature vectors. We then quantify the association between demographic similarity and response similarity by computing a Pearson correlation over all unique participant pairs.

Let $\mathbf{x}^{\text{demo}}_i$ denote the standardized demographic feature vector for participant $i$, and $\mathbf{x}^{\text{resp}}_i$ denote the standardized response vector over evaluation items. We define pairwise similarities as:
\begin{equation}
\small
d_{ij} = \cos\!\left(\mathbf{x}^{\text{demo}}_i, \mathbf{x}^{\text{demo}}_j\right),
\end{equation}
\begin{equation}
\small
s_{ij} = \cos\!\left(\mathbf{x}^{\text{resp}}_i, \mathbf{x}^{\text{resp}}_j\right).
\end{equation}
We then flatten the upper triangle of the similarity matrices and compute the human baseline as:
\begin{equation}
\small
r^{\text{human}} = \mathrm{corr}\!\left(\{d_{ij}\}_{i<j}, \{s_{ij}\}_{i<j}\right),
\end{equation}

Empirically, we obtain
\[
r^{\text{human}}=0.132 \quad (p<0.001),
\]
implying that demographic similarity explains only $r^2\approx 1.5\%$ of the variance in response similarity. This provides a conservative baseline for how much demographic structure is present in \emph{real} responses.

\paragraph{Demographic accentuation in personas:}
For each model $m$ and persona construction case $c$, we compute an analogous coefficient by replacing human response similarity with similarity between the corresponding AI persona answers:
\begin{equation}
    \small
    r^{\text{AI}}_{m,c}=\mathrm{corr}\big(d_{ij},\, s^{\text{AI}}_{m,c,ij}\big).
\end{equation}
Demographic accentuation is the difference from the human baseline and its normalized percentage:

\begin{align}
    \small
    \Delta r_{m,c} &= r^{\text{AI}}_{m,c}-r^{\text{human}}, \\
    \small
    \mathrm{Bias\%}_{m,c} &=100\times \frac{\Delta r_{m,c}}{r^{\text{human}}}.
    \label{eq:bias_percent}
\end{align}
Positive values show persona makes demographically similar individuals \textit{too similar} (over-accentuation), while negative values indicate that the persona \textit{underplays} demographic structure. Our \textit{bias} metric captures demographic accentuation relative to human baselines, not normative unfairness or discrimination.

\paragraph{Results across cases and models:}
The results in Table~\ref{tab:case_gpt4} and Appendix Tables~\ref{tab:full_results_models_a}--\ref{tab:full_results_models_b} show a consistent pattern. First, \textbf{demographic-only prompting strongly over-accentuates demographic similarity}. For GPT-4o, Case~1 yields $\mathrm{Bias\%}=101.23$ (Table~\ref{tab:case_gpt4}), i.e., more than doubling the demographic signal observed in the human baseline. Claude-3.5-Sonnet exhibits an even larger effect in Case~1 ($\mathrm{Bias\%}=115.67$; Table~\ref{tab:full_results_models_a}). This indicates that, when instructed with only demographic variables, models tend to collapse behavior toward demographic stereotypes.

Second, \textbf{adding sociopsychological grounding substantially reduces demographic accentuation}, though the magnitude varies by facet. For GPT-4o, Case~2 (demographics + identity) reduces bias to near-baseline ($\mathrm{Bias\%}=8.30$), whereas Case~3 (demographics + traits) still shows non-trivial over-accentuation ($\mathrm{Bias\%}=75.50$), suggesting that traits alone do not necessarily prevent demographic clustering when demographics remain explicit inputs. In contrast, Case~4 (full conditioning) slightly \emph{under-shoots} the human baseline for GPT-4o ($\mathrm{Bias\%}=-6.40$), indicating that SCOPE can preserve behavioral alignment.

Third, \textbf{non-demographic personas are the most conservative with respect to demographic bias}. For GPT-4o, Case~7c (traits + identity, no demographics) yields $\mathrm{Bias\%}=-56.35$ while maintaining high correlation with human response structure ($\bar{r}=0.658$; Table~\ref{tab:case_gpt4}). A similar pattern appears across other models in Appendix Tables~\ref{tab:full_results_models_a}--\ref{tab:full_results_models_b}, where identity- and traits-driven variants tend to be negative (or substantially reduced) in Bias\% relative to demographic-only variants.


\subsection{Group-based SCOPE Persona Analysis}
\label{subsec:race_analysis}

\begin{table}[t]
\centering
\small
\begin{tabular}{lcc}
\hline
\textbf{Race} & \textbf{N} & \textbf{Baseline ($r$)} \\
\hline
White & 41 & 0.155 \\
African American & 24 & 0.208 \\
Hispanic/Latino & 25 & 0.076 \\
Asian & 18 & -0.004 \\
American Indian/Alaska Native & 9 & -0.005 \\
Other & 4 & N/A \\
Native Hawaiian/Pacific Islander & 3 & N/A \\
\hline
\end{tabular}
\caption{
Race-conditioned human baseline correlations between demographic similarity and response similarity.
Baselines are not reported for groups with insufficient sample size.
}
\label{tab:race_human_baseline}
\end{table}

SCOPE famework also enables demographic-conditioned evaluation, allowing us to examine whether persona construction strategies introduce group-specific overdependencies relative to human behavior. We focus on race in this section (as it shows the highest correlation as shown in Table~\ref{tab:demographic_factor_correlation}), comparing AI demographic response correlations against race-conditioned human baselines to understand if certain persona cases are too dependent on a specific social groups. 

\paragraph{Human baselines differ substantially by race:}
Table~\ref{tab:race_human_baseline} reports the human baseline correlations computed within racial groups. For White ($r=0.155$) and African American or Black participants ($r=0.208$), demographic similarity moderately predicts response similarity, indicating that demographic structure is present in human behavior. In contrast, Hispanic or Latino participants exhibit a weaker baseline, while Asian and American Indian or Alaska Native participants show near-zero correlations. These results indicate that the degree to which demographics explain human responses varies considerably across racial groups and is not a strong indicator of behavior.

\paragraph{Bias is strongly case-dependent and race-dependent:}
Table~\ref{tab:race_ai_bias_combined} shows AI demographic response correlations by race and persona case, together with bias accentuation relative to the human baseline. Across all racial groups, \textbf{Case~1 (Demographics Only)} consistently produces the largest positive bias, indicating that personas constructed solely from demographic attributes substantially over-accentuate demographic structure relative to humans. This effect is particularly pronounced for groups with weak human baselines, such as Asian and American Indian or Alaska Native participants. This indicates possibilities of creating personas that can be racially charged or too flattened to represent a participant from a given group. This also raises challenges of stereotyping in certain cases \cite{cheng2023marked}. It is therefore important to understand what case and persona structure works for a task.

\begin{table*}[t]
\centering
\footnotesize
\begin{tabular}{lccccc}
\hline
\textbf{Persona Case} &
\textbf{White} &
\textbf{African American} &
\textbf{Hisp/Lat} &
\textbf{Asian} &
\textbf{AI / AN} \\
\hline
\textbf{Human baseline} &
0.155 &
0.208 &
0.076 &
$-$0.004 &
$-$0.005 \\

\hline
\textbf{Dem. (\#1)} &
0.446 (+0.291) &
0.449 (+0.242) &
0.252 (+0.176) &
0.216 (+0.220) &
0.250 (+0.255) \\

\textbf{Dem.+Narratives (\#2)} &
0.226 (+0.071) &
0.183 ($-$0.025) &
0.040 ($-$0.036) &
0.096 (+0.100) &
0.237 (+0.241) \\

\textbf{Dem.+Traits (\#3)} &
0.341 (+0.186) &
0.384 (+0.176) &
0.236 (+0.160) &
0.115 (+0.119) &
0.181 (+0.186) \\

\textbf{Full Value (\#4)} &
0.161 (+0.006) &
0.175 ($-$0.033) &
0.065 ($-$0.011) &
0.095 (+0.099) &
0.081 (+0.086) \\

\hline
\textbf{AI Summary (\#5C)} &
0.108 ($-$0.047) &
0.137 ($-$0.071) &
0.037 ($-$0.039) &
0.085 (+0.089) &
0.257 (+0.262) \\

\textbf{Identity Only (\#2D)} &
0.037 ($-$0.118) &
0.057 ($-$0.151) &
$-$0.064 ($-$0.140) &
$-$0.038 ($-$0.034) &
0.192 (+0.197) \\

\hline
\end{tabular}
\caption{
AI demographic response correlations ($r$) by \emph{persona construction case} and race.
The first row reports race-conditioned human baselines.
Values in parentheses denote bias accentuation relative to the corresponding human baseline.
Positive values indicate demographic over-accentuation; negative values indicate reduced demographic dependence.
AI / AN = American Indian or Alaska Native, Hi/Lat = Hispanic or Latino
}
\label{tab:race_ai_bias_combined}
\end{table*}

Incorporating sociopsychological information mitigates this effect. \textbf{Case~2 (Demographics + Narrative)} and \textbf{Case~3 (Demographics + Traits)} reduce bias across most groups. In contrast, \textbf{Case~4 (Full Conditioning)} brings AI behavior close to the human baseline for several races and yields near-zero or negative bias for African American or Black and Hispanic or Latino participants. Ideally we see that \textbf{Demographics + Traits + Augmented} show the most ideal required alignment without stereotyping or misrepresentation. Negative bias can also be a problem where the persona created is too dissimilar to the group in question. Therefore, a balance is needed for the right persona behaviour without stereotyping or misrepresenting a group. 

\begin{table*}[t]
\centering
\footnotesize
\begin{tabular}{lcc}
\hline
\textbf{Case} & \textbf{Group Bias} & \textbf{Std. Dev.} \\
\hline
Case\_1: Demographics Only & +0.237 & 0.043 \\
Case\_2: Demographics + Narratives & +0.070 & 0.113 \\
Case\_3: Demographics + Traits & +0.165 & 0.028 \\
Case\_4: Full Information & +0.030 & 0.059 \\

Case\_5a: AI Summary (300--500w) & -0.042 & 0.084 \\
Case\_5b: AI Summary (500--1000w) & -0.033 & 0.094 \\
Case\_5c: AI Summary (1000--1500w) & +0.039 & 0.139 \\
Case\_5d: AI Summary (No Limit) & -0.037 & 0.121 \\

Case\_7a: Narratives Only & -0.049 & 0.145 \\
Case\_7b: Traits Only & -0.068 & 0.102 \\
Case\_7c: Narratives + Traits & -0.053 & 0.088 \\
\hline
\end{tabular}
\caption{
Average group based bias (in this case Race) accentuation by persona construction strategy across racial groups.
Only officially defined persona cases (Cases~1--5 and Case~7) are reported.
Positive values indicate increased demographic dependence relative to humans, while negative values indicate reduced dependence.
}
\label{tab:avg_bias_by_case}
\end{table*}

\paragraph{Which persona designs increase or reduce bias:}
Table~\ref{tab:avg_bias_by_case} summarizes bias accentuation averaged across racial groups. Demographics-only personas (Case~1 and Case~1S) show the highest average bias, confirming that reliance on demographic attributes alone encourages stereotyping. Hybrid personas that combine demographics with narratives or traits reduce bias but remain positive on average. In contrast, personas that remove explicit demographic information, such as identity-only, traits-only, or augmented non-demographic variants, consistently yield negative bias, indicating reduced demographic dependence relative to humans but also possible misrepresentation of a group. We also see in cases without demographic information, there is complete loss in patterns of behacioural ques leading to negative bias as well. These results show that demographic based group bias in persona-based simulations is neither uniform nor inevitable. Bias emerges most strongly for specific racial groups under specific persona construction strategies, and can be systematically reduced through sociopsychological grounding. We analyze just race in this subsection but each sociodemographic attribute provides new insights on the dependency of the model generating the persona.

\section{Discussion and Conclusion}

Our goal is to re-examine how synthetic personas are constructed, evaluated, and interpreted in NLP/AI. Although personas are widely used for evaluation, and social simulation, our results reveal a systematic mismatch between demographic-centric persona design and the structure of real human behavior, with implications for persona modeling, evaluation, and bias-aware design.

\noindent\textbf{Demographic-centric personas are a structural bottleneck.}
Personas built solely from sociodemographic attributes (Case~1) consistently underperform across models, showing lower behavioral correlation, lower accuracy, and the strongest demographic bias amplification (Tables~\ref{tab:case_gpt4}, \ref{tab:full_results_models_a}, \ref{tab:full_results_models_b}), a pattern that also holds on SimBench. In human data, demographic similarity explains only $\sim$1.5

\noindent\textbf{Sociopsychological grounding improves alignment.}
Adding values, traits, identity narratives, and behavioral signals steadily improves alignment across models, with full conditioning (Case~4) achieving the highest correlation while reducing demographic bias. These gains are not driven by verbosity: AI-generated summaries (Case~5) are often longer yet underperform structured, human-grounded representations, indicating that persona \emph{structure} matters as much as information quantity. Consistently, non-demographic personas based on traits and narrative identity (Case~7c) match or exceed full conditioning in correlation while substantially reducing bias, showing that demographics are neither necessary nor sufficient.

\noindent\textbf{Bias emerges from persona design choices.}
Demographic-only personas systematically over-accentuate demographic clustering, effectively encoding stereotypes into simulated behavior. In contrast, SCOPE reduces bias without sacrificing realism: full conditioning closely matches the human baseline, while traits+narrative personas are the most conservative with respect to demographic amplification while maintaining strong alignment. This supports a practical principle: bias reduction is best achieved by replacing coarse demographic proxies with psychologically meaningful structure, rather than removing structure altogether.

\noindent\textbf{SCOPE as an augmentation framework.}
Rather than replacing existing persona systems, SCOPE functions as a modular augmentation layer. Applied to Nemotron personas, SCOPE-based augmentation improves accuracy and alignment on SimBench, outperforming Nemotron’s personas. This positions SCOPE not merely as a dataset, but as a general-purpose framework for enriching persona representations.

We release only the SCOPE personas, the pipeline code to create augmented SCOPE personas, AI profile materials, and report aggregate statistics to omit any re-identification risk. We release the evaluation code and the SCOPE persona framework in the following Github repository\footnote{https://github.com/SalesforceAIResearch/SCOPE-Persona-Framework/} from which you can generate your own SCOPE augmented personas given a set of seed information. We also release our set of SCOPE persona collection in Hugging Face\footnote{https://huggingface.co/datasets/Salesforce/SCOPE-Persona/}, which consists of 1M synthetic personas that are curated to have additional faceted social and behavioural information. These personas have shown to outperform the existing persona frameworks, especially in behavioural and social tasks, and hence can be used in similar environment settings.


\section*{Ethics Statement}

We collected a two-hour sociopsychological questionnaire from U.S.-based adult participants under review by the institution’s Ethics Office. Participants provided informed consent and were compensated \$50 for completion. To reduce privacy risk, we avoided collecting and removed direct identifiers (e.g., legal names, email addresses, phone numbers, street addresses). Free-text responses were stored under anonymized participant IDs and screened to remove accidental disclosure of sensitive personal information. Because persona conditioning can amplify stereotypes, we measure demographic accentuation (Section~\ref{subsec:bias}) and report persona variants that achieve strong behavioral alignment while reducing demographic clustering. 


The full study protocol underwent internal ethical review prior to deployment. The review examined risks associated with collecting sociopsychological and narrative data, potential re-identification vectors, and the fairness implications of using such data for AI persona modeling. To protect participants, we intentionally avoided collecting direct personal identifiers such as names, email addresses, street addresses, or phone numbers. All responses were stored under anonymized participant identifiers, and raw text responses were screened for accidental disclosure of sensitive personal information. 

No sensitive protected health information or financial identifiers were collected. The anonymized dataset used for modeling contains only aggregated demographic categories and de-identified narrative text. Access to the synthetic dataset is restricted to the research team, and all analyses reported in this paper were conducted exclusively on the de-identified corpus. These procedures were designed to provide strong privacy protections while enabling rigorous sociotechnical research on persona-grounded AI systems.

Importantly, this work does not claim that AI models can replace or faithfully replicate individual human personas. Our evaluation framework is intentionally centered on patterns of behavioral similarity and aggregate tendencies, emphasizing structural alignment and averaged response behavior rather than individual-level substitution. The primary takeaway of this work is methodological: it demonstrates how persona construction choices, particularly the inclusion of sociopsychological structure beyond demographics—substantially affect behavioral realism, bias amplification, and downstream performance. We therefore position SCOPE as guidance on how to build better and more responsible personas, rather than as a claim of human replacement. 

In regards to AI usage in the paper, we used AI-assisted tools solely for grammatical and stylistic corrections of our writing. All scientific content, analysis, and conclusions are the authors’ own.

\section*{Limitations}
Our study acknowledges the following limitations. First, the participant pool is U.S.-based; while we target Census-like proportions and oversample several underrepresented groups, behavioral norms and identity narratives are culturally contingent and may not transfer globally. Second, our evaluation uses a fixed survey instrument; although it spans eight primary facets and multiple response modalities, it cannot cover all drivers of behavior (e.g., longitudinal life events, situational stressors, or fine-grained local context). Third, correlation-based structural evaluation captures relative response patterns but does not guarantee causal fidelity or calibration; two systems can correlate similarly while differing in absolute distributions or rationale quality. Finally, our data-quality controls for AI-authored responses are imperfect; automated detectors are at best heuristic, and filtering decisions may introduce selection artifacts. Our future work intends to incorporate stronger verification (e.g., synchronous interviews, response-time features, or multi-signal provenance checks) and broader cross-cultural replications.

\bibliography{custom}

\appendix
\section{Appendix}
\label{sec:appendix}

This appendix provides additional methodological detail and extended empirical results that support the core claims of the paper but are omitted from the main text for space and clarity. In particular, we present (i) a complete breakdown of the sociopsychological facets underlying SCOPE, (ii) the end-to-end persona construction and evaluation pipeline, and (iii) full model-level results across all persona construction cases. 
\subsection{Facet-level grounding and coverage.}
Table~\ref{tab:scope_facets} details the eight sociopsychological facets that constitute the SCOPE framework, including their behavioral motivation, question counts, and theoretical grounding. As discussed in the main paper, SCOPE explicitly distinguishes between facets used for \emph{persona conditioning} (observable or self-reported attributes such as demographics, personality traits, and identity narratives) and facets reserved for \emph{evaluation} (values, behavioral patterns, professional identity, and creativity).

\begin{table}[t]
\centering
\footnotesize
\begin{tabular}{lcc}
\toprule
\textbf{Question Type} & \textbf{Frequency} & \textbf{\%} \\
\midrule
Likert-scale (1--5) & 51 & 36.17 \\
Dropdown (single-choice) & 37 & 26.24 \\
Multiple-choice & 20 & 14.18 \\
Open-ended (narrative/creative) & 30 & 21.28 \\
Short free-text & 3 & 2.13 \\
\midrule
\textbf{Total} & \textbf{141} & \textbf{100.00} \\
\bottomrule
\end{tabular}
\caption{
Distribution of question types in the SCOPE questionnaire.
Percentages are computed over the full 141-item survey and illustrate the balance between structured (Likert, multiple-choice) and expressive (open-ended) response formats.
}
\label{tab:question_type_distribution}
\end{table}

\begin{table}[t]
\centering
\footnotesize
\begin{tabular}{lcc}
\toprule
\textbf{SCOPE Facet} & \textbf{Freq} & \textbf{\%} \\
\midrule
Demographic Information & 13 & 9.22 \\
Sociodemographic Behavior & 37 & 26.24 \\
Values \& Motivations & 22 & 15.60 \\
Personality Traits (Big Five) & 30 & 21.28 \\
Behavioral Patterns \& Preferences & 13 & 9.22 \\
Identity Narratives & 10 & 7.09 \\
Professional Identity & 10 & 7.09 \\
Creativity \& Innovation & 6 & 4.26 \\
\midrule
\textbf{Total} & \textbf{141} & \textbf{100.00} \\
\bottomrule
\end{tabular}
\caption{
Distribution of questions across the eight sociopsychological facets in the SCOPE questionnaire.
Percentages are computed over the full 141-item instrument.
}
\label{tab:facet_question_distribution}
\end{table}

\begin{figure}[t]
    \centering
    \includegraphics[scale = 0.075]{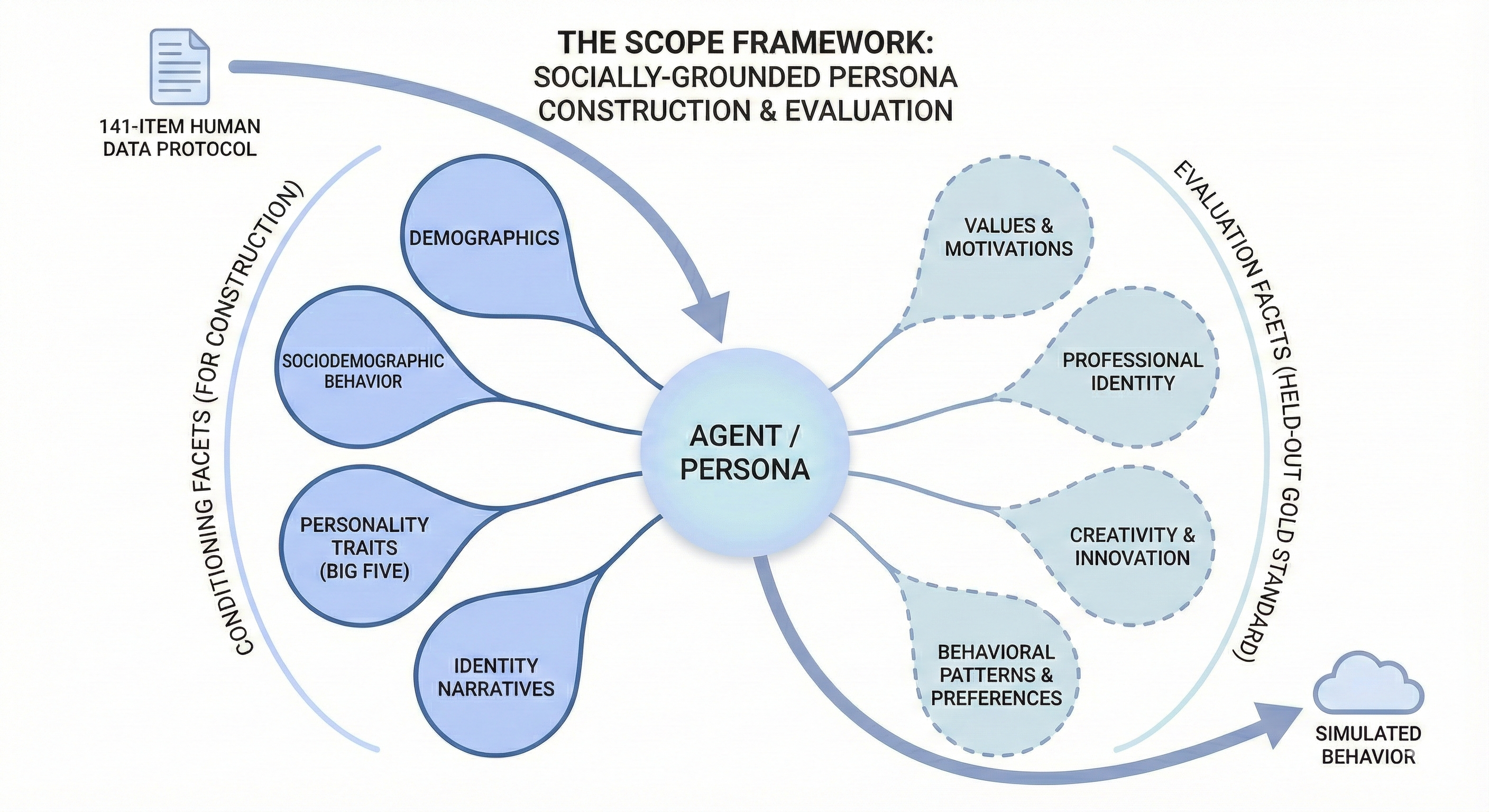}
    \caption{Figure illustrates the facet used for constructing the persona to the ones used for evaluation as a baseline in creating better personas. 
    }
    \label{fig:facet_pipeline}
\end{figure}

\begin{table*}[t]
\centering
\footnotesize
\begin{tabular}{p{2.3cm} p{7.6cm} c p{4cm}}
\hline
\textbf{Facet} &
\textbf{Description} &
\textbf{\#} &
\textbf{Source / References} \\
\hline
\textbf{Demographic Information} &
Captures foundational sociodemographic attributes (e.g., age, gender, race/ethnicity, education, income, political orientation). Although insufficient to explain behavior on their own, such variables ground broader identity, social context, and structural constraints. &
13 &
\textit{Source:} Social Identity (Demographics) \newline
\textit{Refs:} \citet{park2023generative} \\

\textbf{Sociodemographic Behavior} &
Measures group-conditioned behavioral tendencies such as media use, digital access, civic engagement, community belonging, and trust in institutions. These behaviors reflect cultural, structural, and lived-experience influences not recoverable from static demographic labels. This facet draws conceptually on Pew Internet \& Technology surveys, GSS behavioral modules, and WVS civic participation items. &
37 &
\textit{Source:} Sociodemographic Indicators \newline
\textit{Refs:} \citet{park2023generative}, \citet{pewInternet}, \citet{davis1999general}, \citet{inglehart2014wvs} \\

\textbf{Values \& Motivations} &
Derived from established value frameworks (Schwartz), this facet captures benevolence, tradition, achievement, risk-taking, moral orientation, and long-term goals. Values strongly predict social judgments, moral reasoning, preference formation, and life choices. &
22 &
\textit{Source:} Fundamental Human Values \newline
\textit{Refs:} \citet{schwartz1992universals}, \citet{joshi2025improving} \\

\textbf{Personality Traits (Big Five)} &
Encodes stable psychological dispositions, openness, conscientiousness, extraversion, agreeableness, neuroticism, that predict interpersonal behavior, communication style, preference structure, and emotional tendencies. These traits offer robust explanatory power over individual variation. &
30 &
\textit{Source:} Psychological Traits (BFI-2-S) \newline
\textit{Refs:} \citet{costaMcCrae1992}, \citet{joshi2025improving} \\

\textbf{Behavioral Patterns \& Preferences} &
Captures lifestyle habits, emotional responses, motivational strategies, and social tendencies that provide fine-grained behavioral texture. These behaviors emerge from identity, personality, environment, and values, not recoverable from demographic or trait-level information alone. &
13 &
\textit{Source:} Behavioral Patterns \newline
\textit{Refs:} \citet{park2023generative} \\

\textbf{Identity Narratives} &
Short open-ended reflections on personal history, identity formation, relationships, formative events, professional and self-description. Narrative identity theory suggests that people enact behavior through internalized life stories that shape goals, meaning-making, and social interaction. &
10 &
\textit{Source:} Identity Narrative \newline
\textit{Refs:} \citet{mcadams1995}, \citet{park2023generative} \\

\textbf{Professional Identity} &
Represents occupational roles, work habits, domain knowledge, responsibilities, and career motivations. This facet enables personas to generate domain-relevant reasoning and simulate realistic professional behavior. &
10 &
\textit{Source:} Professional Identity \newline
\textit{Refs:} \citet{joshi2025improving} \\

\textbf{Creativity \& Innovation} &
Assesses expressive flexibility, hypothetical reasoning, narrative originality, and creative decision-making. Creative outputs reflect openness, cognitive flexibility, and a persona’s interpretive worldview—providing unique behavioral nuance beyond structured survey items. &
6 &
\textit{Source:} Creativity Tasks \newline
\textit{Refs:} \citet{joshi2025improving}, \citet{cheng2023marked}, \citet{venkit2025tale} \\
\hline
\end{tabular}
\caption{
The eight sociopsychological facets of the SCOPE framework.  
Facets 1--4 form the conditioning inputs used for synthetic persona construction;  
facets 5--8 serve as held-out evaluation dimensions to assess behavioral grounding, generalization, and fidelity.  
Question counts derive from the 141-item persona dataset, and facet categories follow the internal persona taxonomy.
}
\label{tab:scope_facets}
\end{table*}

\subsection{End-to-end Pipeline.}
Figure~\ref{fig:pipeline} visualizes the complete SCOPE workflow, from human data collection to persona construction, evaluation, and aggregation. The pipeline highlights how human-grounded sociopsychological data are serialized into multiple persona variants, evaluated against held-out behavioral targets, and summarized through correlation, accuracy, and bias metrics. 

\subsection{Survey instrument Overview}
\label{sec:appendix_survey_info}
We used a 141-item sociopsychological questionnaire spanning eight facets: Demographic Information (13), Sociodemographic Behavior (37), Values \& Motivations (22), Personality Traits (30), Behavioral Patterns \& Preferences (13), Personal Identity \& Life Narratives (10), Professional Identity \& Career (10), and Creativity \& Innovation (6). The instrument mixes discrete-choice and scale-based items with longer narrative prompts to capture both structured attitudes and expressive self-narratives. For space, we do not reproduce the full questionnaire; instead, Table~\ref{tab:survey_examples_by_facet} provides one illustrative example item per facet along with response formats.

\begin{table*}[t]
\centering
\footnotesize
\begin{tabular}{p{2.7cm} p{5.6cm} p{1.7cm} p{4.2cm}}
\toprule
\textbf{Facet} & \textbf{What it measures} & \textbf{Primary response format(s)} & \textbf{Example item (one per facet)} \\
\midrule

\textbf{Demographic Information} &
Baseline social position and structural context (e.g., age, gender, education, race/ethnicity, political orientation). &
Multiple choice; short text. &
\emph{“Select Your Age”} (10 bins: \textit{Less than 20} … \textit{65 or older}). \\

\textbf{Sociodemographic Behavior} &
Digitally mediated access and civic/media behaviors (internet/AI use, privacy concern, voting importance, institutional trust, cultural belonging). &
Dropdown scales (frequency / agreement); some categorical. &
\emph{“How often do you use AI tools like ChatGPT or Midjourney?”} (frequency scale: \textit{Several times a day} … \textit{Never}). \\

\textbf{Values \& Motivations} &
Motivational priorities and moral/social goals (Schwartz-style value statements; short-/long-term goals). &
Likert (1--5) + multi-select goals. &
\emph{“It’s very important to me to help the people around me. I want to care for others.”} (1=\textit{Very Inaccurate} … 5=\textit{Very Accurate}). \\

\textbf{Personality Traits (Big Five)} &
Stable dispositions across the Big Five (short-form trait statements; mixture of positively and negatively keyed items). &
Likert (1--5). &
\emph{“I am organized. I like to keep things in order.”} (1=\textit{Very Inaccurate} … 5=\textit{Very Accurate}). \\

\textbf{Behavioral Patterns \& Preferences} &
Everyday routines and social/goal regulation (motivation under setbacks, recognition response, weekend routines, social comfort). &
Mixed: narrative (min word count) + multiple choice. &
\emph{“How do you stay motivated when things don’t go as planned?”} (open response; 20+ words). \\

\textbf{Personal Identity \& Life Narratives} &
Narrative identity: autobiographical self-concept, turning points, relationships, neighborhood context, identity change. &
Narrative prompts (min word count). &
\emph{Crossroads prompt: “Was there a moment… where multiple paths were available…? Tell the whole story.”} (open response; 50+ words). \\

\textbf{Professional Identity \& Career} &
Occupational self-concept and work ecology (workday structure, success criteria, tools/platforms, time sinks, AI/automation impacts). &
Narrative prompts (min word count) + one Likert item. &
\emph{“Walk me through a typical workday, highlighting your most critical responsibilities and primary goals.”} (open response; 50+ words). \\

\textbf{Creativity \& Innovation} &
Expressive flexibility and generative ideation (creative pride, hypothetical projects, invented traditions/holidays, metaphorical self). &
Long-form creative prompts (min word count). &
\emph{“Invent a holiday and describe how people would celebrate it.”} (open response; 100+ words). \\

\bottomrule
\end{tabular}
\caption{
Questionnaire summary with one illustrative example item per SCOPE facet. The table is intended to communicate the construct coverage and the response modalities used for modeling and evaluation.
}
\label{tab:survey_examples_by_facet}
\end{table*}

\subsection{Survey Statistics}
\label{sec:appendix_dataset_stats}
Our survey contains 141 questions across eight facets (Table~\ref{tab:scope_facets}). The question-type composition is intentionally mixed to support both structured evaluation and open-ended behavioral analysis: \textbf{51 opinion-scale items (1--5 Likert), 37 dropdown-scale items, 30 multi-format narrative prompts, 20 multiple-choice items, and 3 short-text items.}

Facet-wise, the instrument includes 13 demographic items, 37 sociodemographic-behavior items, 22 values/motivations items, 30 Big Five trait items, 13 behavioral-pattern items, 20 identity prompts (10 identity narrative prompts, 10 professional-identity prompts), and 6 creativity prompts. Table~\ref{tab:question_type_distribution}--\ref{tab:facet_question_distribution} shows the frequency distribution of the types of questions. 

For narrative responses, we report (i) token/word-count distributions by facet (identity, professional, creativity), including median and interquartile range, (ii) missingness rates (items answered as ``N/A'' or left blank), and (iii) per-participant completion time statistics. These summaries are included to support reproducibility and to contextualize model performance on facets with substantially different response entropy and annotation density.

\subsection{Framework Examples}

We show how each of the persona framework differs by showcasing examples of each framework in Table~\ref{tab:persona_examples}. We see how our framework contains more details and information to process behavioural and social tasks better. 

\begin{table*}[t]
\centering
\footnotesize
\begin{tabular}{P{3.2cm} P{8.5 cm} P{3.4cm}}
\hline
\textbf{Persona Paradigm} &
\textbf{Description \& Illustrative Example} &
\textbf{Representative Work} \\
\hline

\textbf{Text-Based Descriptions} &
\textit{Short free-text persona summaries used to condition model behavior.}
\newline
\textbf{Example:} “I am a friendly and curious person who enjoys learning new things and helping others. I value honesty and creativity, and I like discussing technology and everyday life.” &
\citet{ge2024scaling} \\

\textbf{Sociodemographic Inputs} &
\textit{Personas defined as structured demographic attribute vectors, often aligned with census statistics.}
\newline
\textbf{Example:} Age: 34; Gender: Female; Race: Hispanic; Education: Bachelor’s degree; Income: \$50k–\$75k; Region: Southwest U.S. &
\citet{nvidia/Nemotron-Personas-USA} \\

\textbf{Demographic + Identity Signals} &
\textit{Extends demographic personas with explicit self-identity or cultural identification statements.}
\newline
\textbf{Example:} Age: 29; Gender: Male; Race: Asian American.
Identity: “I see myself as both American and strongly connected to my family’s cultural traditions. I value collaboration and community belonging.” &
\citet{lee2025spectrum} \\

\textbf{Interview-Based Personas} &
\textit{Personas derived from interview transcripts or long-form questionnaires capturing lived experience.}
\newline
\textbf{Example:} “I work as a mid-level software engineer maintaining legacy systems. I prefer stability but feel drawn to more creative work. I avoid conflict but speak up on technical issues.” &
\citet{park2023generative} \\

\textbf{SCOPE (Ours)} &
\textit{Personas constructed from structured sociopsychological facets with explicit conditioning and evaluation separation.}
\newline
\textbf{Example:} Demographics (32-year-old Black woman, urban Midwest); Sociodemographic behavior (high civic engagement);
Personality traits (high conscientiousness, moderate openness);
Identity narrative (“I view my career as a way to support my community”).
Evaluation facets (values, behavioral patterns, professional identity, creativity) held out. &
\textit{This work} \\
\hline
\end{tabular}
\caption{
Illustrative persona examples across major persona construction paradigms.
Examples are schematic and intended to highlight differences in representational structure.
}
\label{tab:persona_examples}
\end{table*}

\subsection{Complete model-level results.}
Tables~\ref{tab:full_results_models_a} and~\ref{tab:full_results_models_b} report the full set of experimental results across all seven model families and all persona construction cases. These tables expand on the summarized findings in Section~\ref{sec:evaluation}, showing that the same qualitative trends, demographic over-accentuation, gains from sociopsychological grounding, and the robustness of non-demographic personas, hold consistently across models with very different architectures and training regimes. 

\begin{table}[t]
\centering
\small
\begin{tabular}{lc}
\hline
\textbf{Demographic Factor} & \textbf{Correlation ($r$)} \\
\hline
Race & +0.110 \\
Political Orientation & +0.090 \\
Country of Residence & +0.082 \\
Occupation & +0.072 \\
Relationship Status & +0.046 \\
Income Satisfaction & +0.044 \\
Religion & +0.028 \\
Nationality & +0.027 \\
Income Category & +0.024 \\
Gender & +0.023 \\
Disability / Health Conditions & +0.005 \\
Age & +0.003 \\
Education Level & +0.002 \\
\hline
\end{tabular}
\caption{
Correlation between individual demographic factors and human response similarity of our human baseline dataset. Values reflect Pearson correlation between demographic similarity and response similarity when considering each factor independently.
}
\label{tab:demographic_factor_correlation}
\end{table}

\subsection{External Data Formats: Nemotron Personas and SimBench Instances}
\label{sec:appendix_external_formats}

To contextualize our SimBench evaluation and Nemotron comparisons, we summarize the \emph{data record formats} used by each external resource.

\paragraph{Nemotron-Personas-USA (persona text + structured attributes).}
The NVIDIA \textsc{Nemotron-Personas-USA} \cite{nvidia/Nemotron-Personas-USA} dataset\footnote{\url{https://huggingface.co/datasets/nvidia/Nemotron-Personas-USA}} consists of fully synthetic U.S. personas generated to mirror census-grounded demographic distributions and provide diverse, non-PII persona descriptions.\footnote{The dataset card notes the exclusion of direct identifiers such as names/addresses and describes the generation pipeline via NeMo Data Designer.}
Each row contains (i) multiple \emph{domain-specific persona paragraphs} (e.g., professional, travel, arts), (ii) a \emph{general persona} summary, and (iii) structured demographic and location attributes (e.g., age, sex, education, occupation, state). Concretely, the dataset viewer exposes fields including:
\textit{uuid}, \textit{persona}, \textit{professional\_persona}, \textit{sports\_persona}, \textit{arts\_persona}, \textit{travel\_persona}, \textit{culinary\_persona}, \textit{cultural\_background}, \textit{skills\_and\_expertise} (and list variant),
\textit{hobbies\_and\_interests} (and list variant),
\textit{career\_goals\_and\_ambitions},
along with \textit{sex}, \textit{age}, \textit{marital\_status}, \textit{education\_level}, \textit{bachelors\_field}, \textit{occupation}, \textit{city}, \textit{state}, \textit{zipcode}, \textit{country}.

\paragraph{SimBench (group prompt template + response distribution).}
\textsc{SimBench} \cite{hu2025simbench} is a benchmark of \emph{group-level human response distributions} compiled from multiple social and behavioral datasets into a unified format. Each instance provides:
(i) a \emph{persona/group prompt template} that describes the respondent group (population-level or demographically grouped),
(ii) a \emph{variable map} used to fill template placeholders (empty for population templates),
(iii) the \emph{question text}, and
(iv) an aggregated \emph{human answer distribution} over discrete options. Specifically, the dataset lists primary fields:
\textit{dataset\_name}, \textit{group\_prompt\_template}, \textit{group\_prompt\_variable\_map}, \textit{input\_template},
\textit{human\_answer}, and
\textit{group\_size}.

\paragraph{Examples:}
Nemotron records typically provide multiple topical paragraphs that can be concatenated into a single persona prompt (as done in our Nemotron baseline), e.g.:
\begin{quote}\footnotesize
\textbf{professional\_persona:} \emph{[paragraph describing job, work style, responsibilities]} \\
\textbf{travel\_persona:} \emph{[paragraph describing travel preferences and aspirations]} \\
\textbf{persona:} \emph{[compact overall persona summary]} \\
\textbf{attributes:} \texttt{sex=female, age=28, education\_level=high\_school, occupation=... , state=WI}
\end{quote}
SimBench instances, in contrast, are centered on a group prompt plus a distributional target:
\begin{quote}\footnotesize
\textbf{group\_prompt\_template:} \emph{``You are a 38 year old Amazon Mechanical Turk worker from the United States.''} \\
\textbf{input\_template:} \emph{[question stem + options]} \\
\textbf{human\_answer:} \texttt{\{A:0.04, B:0.08, C:0.03, D:0.81, E:0.04\}}
\end{quote}

\begin{table*}[t]
\centering
\footnotesize
\begin{tabular}{P{2.8cm} P{10.2cm} P{2.6cm}}
\hline
\textbf{SCOPE Case} &
\textbf{Description \& Synthetic Persona Example (Illustrative)} &
\textbf{Included Facets} \\
\hline

\textbf{Case 0: No Persona} &
\textit{Baseline: no persona conditioning text is provided to the model.}
\newline
\textbf{Example:} (No persona preamble; model answers questions with default behavior.) &
None \\

\textbf{Case 1: Demographics Only} &
\textit{Structured demographic attribute vector.}
\newline
\textbf{Example:} 
\newline
\textbf{Select Your Age:} 40--44;
\textbf{Gender:} Male;
\textbf{Education:} Associate degree;
\textbf{Occupation:} Logistics Supervisor;
\textbf{Country:} USA;
\textbf{Race/Ethnicity:} Black;
\textbf{Nationality:} American;
\textbf{Relationship:} Married;
\textbf{Political Orientation:} Moderate;
\textbf{Income Category:} \$50k--\$75k;
\textbf{Religion:} Christian (non-denominational). &
Demographics \\

\textbf{Case 2: Demographics + Identity Narratives} &
\textit{Demographics plus short interview-style identity narratives (personal + professional) that contextualize lived experience.}
\newline
\textbf{Example:}
\newline
\textbf{(Demographics)} 27--29, Female, Bachelor’s, USA, Latina, single, left-leaning.
\newline
\textbf{Professional Identity (snippet):} “I’m a public health coordinator. My day is split between community outreach, planning clinics, and writing reports. I measure success by whether residents actually show up and feel respected.”
\newline
\textbf{Personal Identity (snippet):} “I grew up translating for my parents, which made me independent early. A turning point was choosing a local college so I could support my family while studying.” &
Demographics + Professional Identity + Personal Identity \\

\textbf{Case 3: Demographics + Traits} &
\textit{Demographics plus structured traits items and goal statements (Likert-style), emphasizing motivational priorities.}
\newline
\textbf{Example:}
\newline
\textbf{(Demographics)} 33--39, Nonbinary, Master’s, USA, White, partnered, center-left.
\newline
\textbf{Traits/Goals (snippet):} 
“Helping people around me”=5;
“Being rich”=2;
“Equal opportunities”=5;
“Taking risks/adventures”=3;
“Secure surroundings”=4;
“Thinking up new ideas/being creative”=4.
\newline
\textbf{Short-term goals:} “Pay down debt; build healthier routines; deepen friendships.”
\textbf{Long-term goals:} “Financial stability; meaningful work; more freedom/flexibility.” &
Demographics + Traits/Goals \\

\textbf{Case 4: Full Conditioning} &
\textit{Combined structured persona: demographics + traits/goals + professional identity + personal identity narratives.}
\newline
\textbf{Example:}
\newline
\textbf{(Demographics)} 50--54, Male, Bachelor’s, USA, Asian American, divorced, moderate.
\newline
\textbf{Traits/Goals (snippet):} “Environmental protection”=5; “Tradition”=2; “Achievement/recognition”=3; “Independence”=5.
\newline
\textbf{Professional Identity (snippet):} “I manage IT operations at a mid-sized hospital. The most time-consuming part is incident response and coordinating vendors. I worry about outages and compliance.”
\newline
\textbf{Personal Identity (snippet):} “I moved often as a kid, which made me adaptable. A crossroads moment was leaving a stable job to retrain in tech after my first layoff.” &
Demographics + Traits/Goals + Professional Identity + Personal Identity \\

\textbf{Case 5c: LLM Narrative Summary (1000-1500 words)} &
\textit{Single first-person narrative persona written by the model (compressed, fluent biography-style).}
\newline
\textbf{Example (shortened):} 
“I’m a 38-year-old single father living in the Midwest. I’ve built my career in customer support and learned to stay calm under pressure. I care about fairness and try to treat people with respect, even when politics get heated. Lately, I’ve been thinking about stability, saving more, and finding work that feels meaningful rather than just busy.” &
AI Narrative Only \\

\textbf{Case 6: LLM Facet Completion} &
\textit{Model is given partial inputs and instructed to infer missing facets to complete a persona profile (ablations vary by what is provided).}
\newline
\textbf{Example (given only demographics):} “Age 22--24, Male, some college, USA, Asian.” \textit{Model infers:} plausible values, goals, identity narrative, and work context consistent with the partial seed. &
Varies (partial seed + inferred facets) \\

\textbf{Case 7: Non-Demographic Persona} &
\textit{Persona excludes demographics entirely; conditioning uses non-demographic signals (e.g., identity-only, traits-only, or traits+identity).}
\newline
\textbf{Example (traits+identity only):}
\newline
\textbf{Identity (snippet):} “I see myself as a caretaker in my community, but I’m private about my personal life. I’m drawn to groups where people look out for each other.”
\newline
\textbf{Traits/Goals (snippet):} “Equality”=5; “Loyalty to friends”=5; “Tradition”=2; “Independence”=4; “Creativity”=3.
\newline
\textbf{Long-term goals:} “Build a stable life; contribute to others; keep autonomy.” &
Traits and/or Identity (no demographics) \\

\hline
\end{tabular}
\caption{
SCOPE persona \textbf{case} examples used in our experiments. All examples are \textbf{synthetic} and included to illustrate the \textbf{format and information structure} of each case.
}
\label{tab:scope_case_persona_examples}
\end{table*}

\begin{sidewaystable*}[t]

\centering

\scriptsize

\resizebox{\textwidth}{!}{%

\begin{tabular}{llrrrrrrrrrrrrrrrr}

\toprule

 & & \multicolumn{4}{c}{\textbf{GPT 4o}} & \multicolumn{4}{c}{\textbf{Claude-3.5-Sonnet}} & \multicolumn{4}{c}{\textbf{Gemini 2.0 Flash}} & \multicolumn{4}{c}{\textbf{Gemini-2.5-Pro-Thinking}} \\

\cmidrule(lr){3-6}\cmidrule(lr){7-10}\cmidrule(lr){11-14}\cmidrule(lr){15-18}

\textbf{Case} & \textbf{Description} & $r$ & Acc.\ (\%) & Bias $\Delta$ & Bias (\%) & $r$ & Acc.\ (\%) & Bias $\Delta$ & Bias (\%) & $r$ & Acc.\ (\%) & Bias $\Delta$ & Bias (\%) & $r$ & Acc.\ (\%) & Bias $\Delta$ & Bias (\%) \\

\midrule

Case\_1 & Demographics Only & 0.624 & 35.07 & 0.134 & 101.23 & 0.627 & 36.00 & 0.153 & 115.666 & 0.596 & 35.32 & 0.071 & 54.001 & 0.588 & 29.78 & 0.074 & 55.646 \\

Case\_2 & Demographics + Narratives & 0.642 & 37.74 & 0.011 & 8.298 & 0.656 & 38.56 & 0.012 & 8.791 & 0.609 & 33.58 & -0.101 & -76.79 & 0.616 & 27.58 & -0.016 & -12.245 \\

Case\_3 & Demographics + Traits & 0.656 & 38.60 & 0.100 & 75.499 & 0.650 & 39.60 & 0.055 & 41.287 & 0.600 & 36.71 & 0.012 & 9.087 & 0.614 & 22.56 & -0.042 & -31.64 \\

Case\_4 & Full Information & 0.667 & 39.67 & -0.008 & -6.401 & 0.663 & 39.98 & 0.024 & 17.909 & 0.614 & 35.51 & -0.064 & -48.671 & 0.633 & 25.63 & -0.016 & -11.74 \\

Case\_5a & AI Summary (300–500w) & 0.645 & 37.90 & -0.040 & -30.373 & 0.642 & 37.95 & -0.036 & -27.367 & 0.602 & 34.44 & -0.077 & -58.009 & 0.608 & 30.34 & -0.103 & -78.201 \\

Case\_5b & AI Summary (500–1000w) & 0.645 & 38.12 & -0.039 & -29.80 & 0.639 & 37.74 & -0.058 & -43.601 & 0.591 & 32.98 & -0.103 & -78.321 & 0.618 & 31.74 & -0.114 & -86.013 \\

Case\_5c & AI Summary (1000–1500w) & 0.649 & 38.07 & -0.051 & -38.42 & 0.648 & 38.33 & -0.055 & -41.981 & 0.591 & 31.81 & -0.117 & -88.529 & 0.620 & 32.13 & -0.120 & -90.75 \\

Case\_5d & AI Summary (No Limit) & 0.646 & 38.26 & -0.041 & -31.116 & 0.638 & 37.92 & -0.056 & -42.387 & 0.594 & 32.17 & -0.096 & -72.49 & 0.614 & 32.13 & -0.111 & -84.019 \\

Case\_6a & AI (Given Demography only) & 0.584 & 25.53 & 0.079 & 59.915 & 0.460 & 27.08 & 0.059 & 44.782 & 0.642 & 19.78 & -0.008 & -5.976 & 0.470 & 24.66 & 0.011 & 8.499 \\

Case\_6b & AI (Given Demography + Narratives) & 0.619 & 25.55 & -0.034 & -25.81 & 0.435 & 30.18 & 0.029 & 21.698 & 0.649 & 20.01 & -0.041 & -30.71 & 0.517 & 25.43 & -0.065 & -49.244 \\

Case\_6c & AI (Given Demography + Traits) & 0.606 & 27.86 & 0.016 & 12.249 & 0.462 & 30.15 & 0.009 & 6.797 & 0.681 & 21.32 & -0.025 & -19.097 & 0.496 & 27.30 & -0.066 & -50.171 \\

Case\_6d & AI (Given Identity alone) & 0.572 & 39.72 & -0.062 & -54.69 & 0.588 & 41.36 & -0.054 & -48.15 & 0.544 & 39.29 & -0.087 & -77.65 & 0.578 & 40.61 & -0.094 & -83.38 \\

Case\_6e & AI (Given Traits alone) & 0.588 & 41.65 & -0.080 & -71.43 & 0.592 & 41.45 & -0.071 & -63.02 & 0.544 & 40.49 & -0.101 & -90.04 & 0.577 & 40.37 & -0.099 & -87.90 \\

Case\_6f & AI (Given Identity + Traits) & 0.608 & 43.42 & -0.046 & -40.40 & 0.614 & 43.15 & -0.038 & -33.91 & 0.561 & 40.76 & -0.079 & -70.22 & 0.586 & 42.43 & -0.082 & -72.79 \\

Case\_7a & Identity Only & 0.633 & 37.44 & -0.086 & -65.16 & 0.649 & 38.26 & -0.072 & -54.323 & 0.601 & 35.75 & -0.101 & -76.547 & 0.604 & 30.22 & -0.123 & -93.316 \\

Case\_7b & Traits Only & 0.624 & 37.95 & -0.094 & -71.092 & 0.639 & 38.92 & -0.088 & -66.353 & 0.595 & 36.28 & -0.121 & -91.436 & 0.595 & 34.27 & -0.115 & -87.329 \\

Case\_7c & Traits+Identity & 0.658 & 39.68 & -0.074 & -56.354 & 0.662 & 40.29 & -0.053 & -40.485 & 0.612 & 37.86 & -0.099 & -74.563 & 0.636 & 33.89 & -0.101 & -76.371 \\

AVERAGE & — & 0.636 & 35.53 &  &  & 0.605 & 36.50 &  &  & 0.613 & 31.68 &  &  & 0.588 & 29.12 &  &  \\

\bottomrule

\end{tabular}}

\caption{Complete synthetic persona results (Cases 1--7) for GPT-4o, Claude-3.5-Sonnet, Gemini 2.0 Flash, and Gemini-2.5-Pro-Thinking. We report Pearson correlation with human responses ($r$), answer accuracy (Acc.), and bias accentuation (Bias $\Delta$ and Bias (\%) relative to the human baseline).}

\label{tab:full_results_models_a}

\end{sidewaystable*}

\begin{sidewaystable*}[t]

\centering

\scriptsize

\resizebox{\textwidth}{!}{%

\begin{tabular}{llrrrrrrrrrrrr}

\toprule

 & & \multicolumn{4}{c}{\textbf{DeepSeek R1}} & \multicolumn{4}{c}{\textbf{GPT 5.1}} & \multicolumn{4}{c}{\textbf{Qwen3 480b}} \\

\cmidrule(lr){3-6}\cmidrule(lr){7-10}\cmidrule(lr){11-14}

\textbf{Case} & \textbf{Description} & $r$ & Acc.\ (\%) & Bias $\Delta$ & Bias (\%) & $r$ & Acc.\ (\%) & Bias $\Delta$ & Bias (\%) & $r$ & Acc.\ (\%) & Bias $\Delta$ & Bias (\%) \\

\midrule

Case\_1 & Demographics Only & 0.583 & 31.40 & 0.005 & 0.046 & 0.593 & 29.30 & 0.116 & 1.082 & 0.545 & 21.40 & 0.105 & 0.977 \\

Case\_2 & Demographics + Narrative & 0.609 & 35.40 & -0.022 & -0.207 & 0.615 & 31.80 & 0.035 & 0.322 & 0.572 & 20.90 & -0.020 & -0.190 \\

Case\_3 & Demographics + Traits & 0.602 & 35.10 & -0.011 & -0.107 & 0.591 & 29.10 & 0.050 & 0.469 & 0.553 & 19.10 & -0.033 & -0.304 \\

Case\_4 & Full Information & 0.625 & 37.20 & -0.039 & -0.363 & 0.621 & 32.40 & 0.000 & 0.003 & 0.586 & 21.70 & -0.020 & -0.188 \\

Case\_5a & AI Summary (300–500w) & 0.584 & 34.90 & -0.062 & -0.575 & 0.596 & 33.30 & -0.017 & -0.160 & 0.546 & 22.20 & -0.046 & -0.427 \\

Case\_5b & AI Summary (500–1000w) & 0.594 & 36.00 & -0.060 & -0.563 & 0.599 & 32.70 & -0.015 & -0.144 & 0.540 & 21.90 & -0.029 & -0.274 \\

Case\_5c & AI Summary (1000–1500w) & 0.591 & 36.00 & -0.053 & -0.490 & 0.602 & 32.50 & -0.048 & -0.451 & 0.556 & 22.60 & -0.024 & -0.221 \\

Case\_5d & AI Summary (No Limit) & 0.589 & 34.70 & -0.071 & -0.658 & 0.597 & 31.70 & -0.048 & -0.442 & 0.552 & 22.00 & -0.057 & -0.533 \\

Case\_6a & AI (Given Demography only) & 0.539 & 23.61 & 0.039 & 29.486 & 0.543 & 21.28 & 0.032 & 24.124 & 0.502 & 14.47 & 0.037 & 28.145 \\

Case\_6b & AI (Given Demography + Identity) & 0.572 & 23.63 & -0.030 & -23.118 & 0.576 & 21.29 & -0.025 & -18.915 & 0.532 & 14.48 & -0.029 & -22.067 \\

Case\_6c & AI (Given Demography + Traits) & 0.560 & 25.76 & -0.018 & -13.811 & 0.564 & 23.22 & -0.015 & -11.300 & 0.521 & 15.79 & -0.017 & -13.183 \\

Case\_6d & AI (Given Narrative alone) & 0.528 & 36.73 & -0.082 & -72.564 & 0.532 & 33.10 & -0.067 & -59.371 & 0.492 & 22.51 & -0.078 & -69.266 \\

Case\_6e & AI (Given Traits alone) & 0.543 & 38.52 & -0.097 & -85.907 & 0.547 & 34.71 & -0.079 & -70.288 & 0.506 & 23.61 & -0.092 & -82.002 \\

Case\_6f & AI (Given Identity + Traits) & 0.561 & 40.15 & -0.067 & -59.763 & 0.566 & 36.19 & -0.055 & -48.897 & 0.523 & 24.61 & -0.064 & -57.046 \\

Case\_7a & Narrative Only & 0.584 & 34.62 & -0.105 & -79.570 & 0.589 & 31.20 & -0.086 & -65.103 & 0.544 & 21.22 & -0.100 & -75.953 \\

Case\_7b & Traits Only & 0.576 & 35.10 & -0.115 & -86.958 & 0.581 & 31.63 & -0.094 & -71.147 & 0.537 & 21.51 & -0.110 & -83.005 \\

Case\_7c & Traits+Identity & 0.608 & 36.70 & -0.090 & -68.138 & 0.612 & 33.07 & -0.074 & -55.749 & 0.566 & 22.49 & -0.086 & -65.040 \\

AVERAGE & — & 0.579 & 33.85 &  &  & 0.584 & 30.50 &  &  & 0.540 & 20.73 &  &  \\

\bottomrule

\end{tabular}}

\caption{Complete synthetic persona results (Cases 1--7) for DeepSeek R1, GPT 5.1, and Qwen3 480b. Metrics are defined as in Table~\ref{tab:full_results_models_a}.}

\label{tab:full_results_models_b}

\end{sidewaystable*}

\section{Analysis of Open-ended and Writing-style based Questions}
\label{sec:appendix_creativity}

In addition to structured multiple-choice and Likert-style evaluation, SCOPE includes a small set of open-ended prompts intended to elicit narrative expression and creative reasoning (Facet 8: Creativity \& Innovation). These prompts are qualitatively important because they test whether persona conditioning preserves \emph{style and narrative structure} rather than only discrete response choices. Following prior parameterized creativity evaluation work \citep{venkit2025tale, chakrabarty2024art}, we quantify differences between human-authored and persona-conditioned LLM narratives along four computational axes inspired by Torrance-style creativity dimensions (flexibility, originality, elaboration, and coherence).

\subsection{Creativity metric definitions}
\label{sec:creativity_metrics}

Let a group of responses (e.g., a persona case, model, or question group) contain $n$ narratives, and let $e_i$ denote the Sentence-BERT embedding for narrative $i$.

\paragraph{Semantic Diversity (Flexibility).}
We measure within-group thematic breadth as the average pairwise semantic distance:
\begin{equation}
\small
\mathrm{Diversity} =
\frac{2}{n(n-1)}
\sum_{i=1}^{n-1}\sum_{j=i+1}^{n}
\left(1 - \cos(e_i, e_j)\right).
\end{equation}
Higher values indicate broader variation in themes and expression.

\paragraph{Semantic Novelty (Originality).}
We measure how much a group's average internal distance deviates from the corpus norm:
\begin{equation}
\small
\mathrm{Novelty} = 2 \times \left| d_{\text{group}} - d_{\text{corpus}} \right|,
\end{equation}
where $d_{\text{group}}$ is the group's average pairwise semantic distance and $d_{\text{corpus}}$ is the corpus-wide average distance. Higher values indicate greater deviation from expected narrative patterns.

\paragraph{Semantic Complexity (Elaboration).}
We operationalize elaboration via a composite of lexical rarity (TF-IDF) and semantic spread (Word2Vec):
\begin{equation}
\small
\mathrm{C}(s) =
0.5 \times \frac{\mathrm{C}_{\text{TFIDF}}(s)}{\max(\mathrm{C}_{\text{TFIDF}})}
+
0.5 \times \frac{\mathrm{C}_{\text{W2V}}(s)}{\max(\mathrm{C}_{\text{W2V}})}.
\end{equation}
Higher values indicate more intricate narratives in both vocabulary and concept dispersion.

\paragraph{Surprisal (Narrative Coherence).}
We measure narrative coherence by the average semantic distance between consecutive sentence embeddings within each response:
\begin{equation}
\small
\mathrm{Surprisal} =
\frac{2}{n-1}
\sum_{i=2}^{n}
\left(1 - \cos(e_{i-1}, e_i)\right),
\end{equation}
where here $e_i$ denotes the embedding of the $i$-th sentence. Lower surprisal corresponds to smoother semantic progression (more coherent flow).

\subsection{Experimental setup}
\label{sec:creativity_setup}

We evaluate creativity on six open-ended SCOPE prompts (Facet 8), comparing human-written responses to LLM-generated responses conditioned on the same persona construction cases used in the main study. We report aggregate comparisons (Human vs.\ AI) and case-level patterns using the four metrics above.

\subsection{Human vs.\ AI: aggregate findings}
\label{sec:creativity_human_vs_ai}

Table~\ref{tab:creativity_human_vs_ai} summarizes aggregate creativity gaps. Consistent with the broader theme of this paper, \emph{persona structure affects behavioral outputs}, we observe that open-ended narrative behavior differs from structured answering behavior: LLM outputs tend to be more elaborative and more deviant from the corpus norm, but less varied across personas and less coherent in narrative flow.

\begin{table*}[t]
\centering
\small
\begin{tabular}{lccc}
\toprule
\textbf{Metric} & \textbf{Human} & \textbf{AI Persona} & \textbf{Direction} \\
\midrule
Semantic Complexity $\uparrow$ & 0.6933 & 0.7924 & AI higher (more elaboration) \\
Surprisal (Coherence) $\downarrow$ & 0.9571 & 1.1384 & AI worse (more semantic jumps) \\
Semantic Diversity $\uparrow$ & 0.1910 & 0.1685 & Human higher (more variation) \\
Semantic Novelty $\uparrow$ & 0.0181 & 0.0329 & AI higher (more deviation) \\
\bottomrule
\end{tabular}
\caption{
Aggregate creativity and narrative-style comparison between human-authored responses and persona-conditioned across all 7 model responses across the creativity prompts. Arrows indicate preferred direction for \textit{human-like} narrative quality: higher diversity/complexity/novelty, and lower surprisal.
}
\label{tab:creativity_human_vs_ai}
\end{table*}

\subsection{Persona-case effects: a creativity–constraint trade-off}
\label{sec:creativity_case_effects}
To understand how persona construction affects narrative-style realism, we compare persona cases by measuring distance from human creativity patterns across the four normalized metrics (equal weight). Interestingly, the most human-like creativity behavior is achieved by \textbf{Case 1 (Demographics Only)} (Table~\ref{tab:creativity_case_rank}), suggesting a “creativity–constraint trade-off”: richer persona scaffolds that improve structured behavioral fidelity (Sections~\ref{sec:evaluation}--\ref{subsec:bias}) may over-constrain open-ended narrative generation and reduce natural variation. This complements our main finding that persona quality is facet-dependent: the optimal conditioning strategy for \emph{structured behavioral prediction} is not necessarily optimal for \emph{open-ended expressive tasks}.

\begin{table*}[t]
\centering
\small
\begin{tabular}{l l c}
\toprule
\textbf{Rank} & \textbf{Case} & \textbf{Distance to Human (lower is better)} \\
\midrule
1 & Case\_1 (Demographics only) & 0.5473 \\
2 & Case\_2D (Narrative only; no demographics) & 0.5763 \\
3 & Case\_2 (Demographics + identity) & 0.5868 \\
4 & Case\_4 (Full conditioning) & 0.6010 \\
5 & Case\_2S (Narrative augmented) & 0.6181 \\
\bottomrule
\end{tabular}
\caption{
Top creativity-aligned persona cases under a normalized, equal-weight distance over semantic complexity, surprisal, semantic diversity, and semantic novelty across all 7 models. Lower distance indicates closer alignment to human creativity patterns.
}
\label{tab:creativity_case_rank}
\end{table*}

\paragraph{Interpretation.}
These results indicate that (i) \textbf{demographic-only personas} can preserve a surprisingly human-like balance of novelty and diversity in unconstrained narrative tasks, while (ii) \textbf{richer sociopsychological scaffolds} (full conditioning and long summaries) tend to increase elaboration but may compress persona-to-persona variability, reducing diversity. Practically, this suggests a deployment guideline: for \emph{creative, narrative, or story-style user simulation}, minimal persona constraints may produce more human-like stylistic variation; for \emph{structured behavior prediction and bias-sensitive simulation}, multi-facet grounding remains preferable (Sections~\ref{sec:evaluation}--\ref{subsec:bias}).

\end{document}